\documentclass[11pt,twoside]{article} 
\usepackage{acta-info}
\usepackage{euler}
\usepackage{algorithmic}
\usepackage{algorithm}
\usepackage{amssymb}

\usepackage{graphicx, epstopdf}
\usepackage{amsmath}

\newcommand{\vol}{\textbf{4,} 1 (2012) 48--66}   
\newcommand{\amss}{\amssubj{%
90C27
}}     
\newcommand{\CRs}{\CRsubj{%
G.1.6
}}   
\newcommand{\keyw}{\keywords{scheduling theory, artificial neural networks, Hopfield neural network, total weighted tardiness problem, quadratic optimization}
}

\begin{document}

\title{A novel Hopfield neural network approach for minimizing total weighted tardiness of jobs scheduled on identical machines}

\maketitle

\twoauthors{
Norbert FOGARASI
}{\href{http://www.hit.bme.hu/}{Department of Telecommunications};
  \href{http://www.bme.hu/}{Budapest University of Technology and Economics}
  Budapest, Hungary;
}{%
\href{mailto:fogarasi@hit.bme.hu}{fogarasi@hit.bme.hu} 
}{%
\href{http://users.itk.ppke.hu/~kami}{K\'alm\'an TORNAI} 
}{\href{http://www.itk.ppke.hu}{Faculty of Information Technology};
  \href{http://www.ppke.hu}{Pazmany Peter Catholic University}
  Budapest, Hungary;
}{%
\href{mailto:tornai.kalman@itk.ppke.hu}{tornai.kalman@itk.ppke.hu} 
}

\oneauthor{
\href{http://neural.hit.bme.hu/levendov/}{J\'anos LEVENDOVSZKY} 
}{\href{http://www.hit.bme.hu/}{Department of Telecommunications};
  \href{http://www.bme.hu/}{Budapest University of Technology and Economics}
  Budapest, Hungary;
}{%
\href{mailto:levendov@hit.bme.hu}{levendov@hit.bme.hu} 
}

\short{N. Fogarasi, K. Tornai, J. Levendovszky}{HNN approach for minimizing TWT on identical machines}

\begin{abstract}
This paper explores fast, polynomial time heuristic approximate solutions to the NP-hard problem of scheduling jobs on N identical machines. The jobs are independent and are allowed to be stopped and restarted on another machine at a later time. They have well-defined deadlines, and relative priorities quantified by non-negative real weights. The objective is to find schedules which minimize the total weighted tardiness (TWT) of all jobs. We show how this problem can be mapped into quadratic form and present a polynomial time heuristic solution based on the Hopfield Neural Network (HNN) approach. It is demonstrated, through the results of extensive numerical simulations, that this solution outperforms other popular heuristic methods. The proposed heuristic is both theoretically and empirically shown to be scalable to large problem sizes (over 100 jobs to be scheduled), which makes it applicable to grid computing scheduling, arising in fields such as computational biology, chemistry and finance.
\end{abstract}

\section{Introduction}
\label{sec:1Intro}
With the advent of grid computing, the classical science of scheduling theory has gained a new sphere of applications. Methods which were originally developed for decision making around limited resources in manufacturing and service industries have been adopted in the areas of computer science, telecommunication and other computationally intensive disciplines such as computational biology, chemistry and finance \cite{Brucker2007, Pinedo2008}. Specifically, in the area of computational finance, where this paper sources its motivation, the problems of portfolio selection, pricing and hedging of complex financial instruments requires an enormous amount of computational resources whose optimal usage is of utmost importance to investment banks. The prices and risk sensitivity measures of complex portfolios need to be reevaluated daily, for which an overnight batch of calculations is scheduled and performed for millions of financial transactions, utilizing thousands of computing nodes. Each job has a well-defined priority and required completion time for availability of the resulting figures to the trading desk, risk managers and regulators. The jobs can generally be stopped and resumed at a later point on a different machine which is referred to as \emph{preemption} in scheduling theory. For simplicity of modeling the problem, machines are generally assumed to be \emph{identical} and there is a known, constant number of machines available. 

The problem of finding optimal schedules for jobs running on identical machines has been extensively studied over the last three decades. Sahni \cite{Sahni1979} presents an $\textit{O}\left( n \log m n \right)$ algorithm to construct a \emph{feasible} schedule, one that meets all deadlines, if one exists, for $n$ jobs and $m$ machines. The basic idea of the algorithm is to schedule jobs with earliest due dates first, but fill up machines with smaller jobs if possible. Note that this method allows the development of an algorithm to compute the minimal amount of unit capacity for which a feasible schedule exists.  This result has been extended to machines with identical functionality but different processing speed, termed \emph{uniform machines}, and jobs with both starting times and deadlines \cite{Martel1982}. However, the scheduling task becomes more difficult when a feasible schedule does not exist and the goal is to minimize some measure of delinquency, often termed \emph{tardiness}. Tardiness of an individual job under a given schedule is defined as the amount of time by which the job finishes after its prescribed deadline, and is considered to be zero if the job finishes on or before the deadline.

In case of minimizing the maximum tardiness across all jobs, Lawler \cite{Lawler1981} shows that the problem is solvable in polynomial time, even with some precedence constraints. Martel \cite{Martel1982} also used his construction to create a polynomial time algorithm to find the schedule which minimizes maximum lateness. However, if our measure concerns the total tardiness instead of the maximal one, then even the single machine, total tardiness problem (without weights) was proven to be NP-hard by Du et al \cite{Du1990}. A pseudopolynomial algorithm has been developed by Lawler \cite{Lawler1977} for this problem, using dynamic programming, but this is for the 1-machine problem and does not have good practical runtime characteristics.

In practical applications, jobs often have relative priorities associated with them, represented by positive real \emph{weights} and the objective becomes minimizing the total weighted tardiness (TWT).
Once the NP-hardness of the TWT problem was established, most of the research work on the problem concerned the development of fast, heuristic algorithms. Dogramaci et al. \cite{Dogramaci1979} propose a simple heuristic for the total (non-weighted) tardiness problem without preemption. Rachamadugu et al. \cite{Rachamadugu1983} then studied the identical machine, total weighted tardiness problem without preemption. They proposed a myopic heuristic and compared this to earliest due date (EDD), weighted shortest processing time (WSPT) and Montagne’s rule on small problem sizes (2 or 5 jobs in total). Azizoglu et al. \cite{Azizoglu1998} worked on an algorithm to find optimal schedule for the unweighted total tardiness problem without preemption, but their branch and bound exponential algorithm is too slow, in practice, for problems with more than 15 jobs. Armentano et al. \cite{Armentano2000} examined the non-weighted problem without preemption, and starting from the KPM heuristic of Koulamas \cite{Koulamas1994} improved upon it, using tabu search. Guinet \cite{Guinet1995} applies simulated annealing to solve the problem with uniform and identical machines and a lower bound is presented in order to evaluate the performance of the proposed method. More recently, Sen et al. \cite{Sen2003} surveyed the existing heuristic algorithms for the single-machine total tardiness and total weighted tardiness problems while Biskup et al. \cite{Biskup2008} did this for the identical machines total tardiness problem and also proposed a new heuristic. Akyol et al. \cite{Akyol2008} provide an excellent recent review of artificial neural network based approaches to scheduling problems and proposes a coupled gradient network to solve the weighted earliness plus tardiness problem on multiple machines. The feasibility of the method is illustrated on a single 8-job scheduling problem.

We suggest a novel heuristic for the TWT problem, based on the Hopfield Neural Network approach which is shown to perform better than existing simple heuristics and has desirable scaling characteristics. Maheswaran et al. \cite{Maheswaran2004} applied a similar approach to the single machine TWT problem and their results were encouraging for a specific 10-job problem.

In this paper, we first map the problem into quadratic optimization and then the Hopfield net is used to provide fast polynomial time, heuristic solution. The topics are organized as follows: in Section 2, we present the problem formulation and the model used, in Section 3 existing heuristics are defined and explained, in Section 4 our novel HNN approach is introduced, in Section 5 numerical results are presented and finally in Section 6 some conclusions are drawn and directions for future research are outlined.

\section{Problem formulation}
\label{sec:2ProbForm}
In this section, we give a formal presentation for the problem of optimally scheduling jobs on finite number of identical processors under constraints on the completion times.  The basic formalism is the following:
\begin{itemize}
\item Given $N$ jobs with sizes ${\bf{x}} = \left\{ {{x_1},{x_2},{x_3}, \ldots {x_N}} \right\} \in {\mathbb{N}^N}$. The processing of the jobs can be stopped and resumed at any time, so the processing time units of each job need not be contiguous. In the literature this condition is known as preemption and also assumes a task started on one machine can continue on another \cite{Brucker2007}.
\item For each job a cutoff time is prescribed by ${\bf{K}} = \left\{ {{K_1},{K_2},{K_3}, \ldots ,{K_N}} \right\} \in {\mathbb{N}^N}$. This constraint defines the time within which the job is to be completed.
\item The constant number of processors, the capacity of the system is denoted by $V \in \mathbb{N}$.
\item We are also given a vector ${\bf{w}} = \left\{ {{w_1},{w_2},{w_3}, \ldots ,{w_N}} \right\} \in {\mathbb{R}^N}{\rm{, }}{w_i} \ge 0,\forall i = 1,\ldots,N$ denoting the relative priority (or weight) of each job, which can be used in the definition of the objective function. 
\end{itemize}

A schedule is represented by a binary matrix ${\bf{C}} \in {\left\{ {0,1} \right\}^{N \times L}}$ where ${C_{i,j}} = 1$ if job $i$ is being processed at time slot $j$, and $L$ denotes the length of the schedule. An example is given in (\ref{eq:e21}) where the parameters are the following: $V = 2$, $N = 3$, ${\bf{x}} = \left\{ {2,3,1} \right\}$, ${\bf{K}} = \left\{ {3,3,3} \right\}$.
\begin{equation}
{\bf{C}} = \left( {\begin{array}{ccc}
1 & 0 & 1 \\
1 & 1 & 1 \\
0 & 1 & 0
\end{array}} \right)
\label{eq:e21}
\end{equation}
The first row in (\ref{eq:e21}) denotes the fact that under this schedule $\mathbf C$, the first job is processed in time steps $1$ and $3$ (note that preemption is used as the processing of this job is not continuous) and therefore the prescribed size $2$ of this job completes within the prescribed cutoff time of time step $3$. Similarly, the $3$ units of the second job complete within the cutoff time of $3$ and the third job is completed ahead of the cutoff time on time step $2$. Summing the columns of matrix $\mathbf C$, we see that the maximal capacity of $V=2$ is fully utilized on each time step.

In order to evaluate the effectiveness of a given schedule $\mathbf C$, we define tardiness of a job as follows:
\begin{equation}
{T_i} = \max \left( {0,{F_i} - {K_i}} \right),
\label{eq:e22}
\end{equation}
where $F_i$ is the actual finish time of job $i$ under schedule $\mathbf C$: ${F_i} = \mathop {\arg \max }\limits_j \left\{ {{C_{i,j}} = 1} \right\}$  (The position of the last $1$ in the $i$th row in scheduling matrix $\mathbf C$.)

The problem can now be stated formally as follows:
\begin{equation}
{{\bf{C}}_{opt}}: = \mathop {\arg \min }\limits_{\bf{C}} \sum\limits_{i = 1}^N {{w_i}{T_i}}.
\label{eq:e23}
\end{equation}
Under the following constraints:
\begin{itemize}
\item The sizes of the scheduled jobs in the scheduling matrix are equal to the predefined amounts:
\begin{equation}
\sum\limits_{j = 1}^L {{C_{i,j}}}  = {x_i},\forall i = 1,\ldots ,N .
\label{eq:e24}
\end{equation}
\item The number of scheduled jobs at any given time instant does not exceed the capacity of the system:
\begin{equation}
\sum\limits_{i = 1}^N {{C_{i.j}}}  \le V,\forall j = 1,\ldots ,L .
\label{eq:e25}
\end{equation}
\end{itemize}

We revisit our previous example with a minor change: $V = 2$, $N = 3$, ${\bf{x}} = \left\{ {2,3,2} \right\}$, ${\bf{K}} = \left\{ {3,3,3} \right\}$ and a weight vector ${\bf{w}} = \left\{ {3,2,1} \right\}$. It can be observed that there is no solution, in which all jobs are completed before their cutoff times. A minimal weighted tardiness solution is the following:
\[
{\bf{C}} = \left( {\begin{array}{cccc}
1 & 0 & 1 & 0 \\
1 & 1 & 1 & 0 \\
0 & 1 & 0 & 1
\end{array}} \right) .
\]

\section{Existing heuristic methods}
\label{sec:3ExistingHeuristics}
Given the NP-hardness of the scheduling task as shown by Du et al. \cite{Du1990}, and therefore the amount of time it would take to find the exact, optimal solution, in most real-world settings the pragmatic approach of finding a fast, sub-optimal, but good solution is followed. As outlined in the introduction, this has been a very active field of research, and most studies use the Earliest Due Date (EDD) and Weighted Shortest Process Time (WSPT) heuristics as benchmarks for evaluating the proposed algorithms. In addition to these, we also outline the recently developed Load Balancing Scheduling (LBS) heuristic and a newly proposed Largest Weighted Process First (LWPF) heuristic. Furthermore, we also outline a random processing method which is used as a low benchmark for our testing results.

\subsection{Earliest due date (EDD) heuristic}
\label{sec:31EDD}

The EDD heuristic orders the sequence of jobs to be executed from the job with the earliest due date to the job with the latest due date. Using the notation of Section 2, we relabel the job indices so that the following inequality holds:
\[
{K_1} \le {K_2} \le \ldots \le {K_N}.
\]
Once this ordering is determined, the jobs are allocated to the machines in this order, always utilizing the maximum available capacity. Once a job finishes and capacity is freed up, the next job using the above ordering is scheduled on the freed up machine. It has been shown in \cite{Pinedo2008} that EDD finds the optimal schedule when one wants to minimize the maximum tardiness on a single machine. However, we note that when the objective function includes the relative priorities of jobs, EDD is at a severe disadvantage, as it does not include consideration of the weights.

\subsection{Weighted shortest processing time (WSPT) heuristic}
\label{sec:32WSPT}
The WSPT method is analogous to the EDD method in that it orders the jobs according to well-defined criteria and then schedules the jobs according to this ordering, utilizing the maximum available 	capacity available. Once a job finishes and capacity is freed up, the next job using the ordering is scheduled. Using the notation of Section 2, the jobs are ordered such that the below holds:
\[
\frac{{{x_1}}}{{{w_1}}} \le \frac{{{x_2}}}{{{w_2}}} \le \ldots \le \frac{{{x_N}}}{{{w_N}}}.
\]

\subsection{Largest weighted process first (LWPF) heuristic}
\label{sec:33LWPF}
This heuristic is analogous to the EDD and the WSPT heuristics with the jobs ordered according to the following inequality:
\[
{w_1} \ge {w_2} \ge \ldots \ge {w_N}.
\]
This is a simple heuristic which allows us get a sense of the importance of considering the weights versus the cutoff times. We have not seen this simple heuristic mentioned anywhere in the literature, but it turns out to have quite good empirical characteristics which is one of the many contributions of this paper.

\subsection{Load Balancing Scheduling (LBS) algorithm}
\label{sec:34LBS}
Laszlo et al \cite{Laszlo2011} suggest a novel deterministic, polynomial time algorithm for scheduling jobs with cutoff times, in the context of load balancing for wireless sensor networks. The basic idea of the algorithm is to start scheduling the jobs backwards from the maximal cutoff time, in order of decreasing cutoff times, always ensuring full capacity is utilized. For a more formal definition of the algorithm, please see \cite{Laszlo2011}. We note that whilst the algorithm has proven to be extremely successful for load balancing, it does not take into account the weights of the jobs and therefore will suffer from the same shortcomings in our setting as EDD.

\subsection{Random method}
\label{sec:35Random}
In this method, the jobs are ordered randomly and are scheduled in such a way as to always fully utilize the maximum available capacity. This unsophisticated heuristic is useful as a low benchmark by repeating it a number of times and taking the best solution over the multiple repeats.

\section{Novel Hopfield neural network (HNN) based approach}
\label{sec:4HNN}
Since the number of possible binary matrices grows exponentially with the number of nodes and the length of the schedule, exhaustive search with complexity ${\textit O}\left( {{2^{N \cdot L}}} \right)$ is generally computationally infeasible to solve the optimization problem at hand. Thus, our goal is to develop a polynomial time approximate solution by mapping the problem to an analogous quadratic optimization problem, similar to the approach of Levendovszky et al \cite{Levendovszky2011} and Treplan et al \cite{Treplan2011}.

We first review the methods available for optimizing equations in quadratic form. Let us assume that matrix $\mathbf W$ is a symmetric matrix of size $n \times n$ and vector $\mathbf b$ is of length $n$. We seek the optimal $n$ dimensional vector $\mathbf y$ which minimizes the following quadratic function \cite{Nocedal2006}:
\begin{equation}
f\left( {\bf{y}} \right) =  - \frac{1}{2}{{\bf{y}}^T}{\bf{Wy}} + {{\bf{b}}^T}{\bf{y}},
\label{eq:e41}
\end{equation}
subject to one or more constraints of the form of
\[
\begin{array}{l}
{\bf{Ay}} \le {\bf{v}} ,\\
{\bf{By}} = {\bf{u}} .
\end{array}
\]
If the problem at hand contains only linear constraints then it can be solved as presented by Murty et al \cite{Murty1988}. In other cases, if the matrix $\mathbf W$ is positive definite, then the function $f\left(\mathbf{y}\right)$ is convex and the problem can be solved with the ellipsoid method presented by Zhi-Quan et al \cite{Zhi-quan2010}. When $\mathbf{W}$ is indefinite, the problem is NP-hard (for details see \cite{Sahni1974}).

A frequently used powerful, heuristic algorithm to solve quadratic optimization problems is the Hopfield Neural Network (HNN). This neural network is described by the following state transition rule:

\[
{{\bf{y}}_i}(k + 1) = {\mathrm{sgn}}\left( {\sum\limits_{j = 1}^N {{{\hat W}_{ij}}} {y_j}(k) - {{\hat b}_i}} \right), i = \mathrm{mod}_N k,
\]
where
\[
\begin{array}{c}
{\bf{d}} =  - {\rm{diag}}\left( {\bf{W}} \right),\\
\widehat {\bf{W}} =  - {\bf{W}} - {\rm{diag}}\left( {\bf{d}} \right),\\
\widehat {\bf{b}} = {\bf{b}} - \frac{1}{2}{\bf{d}}.
\end{array}
\]
Using the Lyapunov method, Hopfield \cite{Hopfield1982} proved that the HNN converges to its fixed-point, as a consequence the HNN minimizes a quadratic Lyapunov function:
\[
{\cal L}({\bf{y}}): =  - \frac{1}{2}\sum\limits_{i = 1}^N {\sum\limits_{j = 1}^N {{{\hat W}_{ij}}} } {y_i}{y_j} + \sum\limits_{i = 1}^N {{y_i}} {\hat b_i} =  - \frac{1}{2}{{\bf{y}}^T}{\bf{\hat Wy}} + {{\bf{\hat b}}^T}{\bf{y}}.
\]
Thus, the HNN is able to solve combinatorial optimization problems in polynomial time under special conditions as it has been demonstrated by Mandziuk \cite{Mandziuk1992, Mandziuk1996}. Using this method in practice we have to handle the following problem: in some cases the HNN method may get stuck in local minimal point of its Lyapunov function.

These methods motivate us to map the existing optimization problem into quadratic form. First the binary scheduling matrix $\mathbf{C}$ is mapped into a binary column vector $\mathbf{y}$ as follows:
\[{\bf{C}} = \left( {\begin{array}{*{20}{c}}
{{C_{1,1}}}&{{C_{1,2}}}& \cdots &{{C_{1,L}}}\\
{{C_{2,1}}}&{{C_{2,2}}}& \cdots &{{C_{2,L}}}\\
 \vdots & \vdots & \ddots & \vdots \\
{{C_{N,1}}}&{{C_{N,2}}}& \cdots &{{C_{N,L}}}
\end{array}} \right) \to \]
\[
{\bf{y}} = {\left( {{C_{1,1}},{C_{1,2}}, \cdots ,{C_{1,L}},{C_{2,1}}, \cdots ,{C_{2,L}},{C_{N,1}}, \cdots ,{C_{N,L}}} \right)^T}.
\]
The original objective function (\ref{eq:e23}) is elaborated as follows:
\[
\min_\mathbf{C} \sum\limits_{i = 1}^N {{w_i}\left( {\max \left( {0,\mathop {\arg \max }\limits_j \left\{ {{C_{i,j}} = 1} \right\} - {K_i}} \right)} \right)} .
\]
This objective is equivalent to:
\[
\min_\mathbf{C} \sum\limits_{i = 1}^N {{w_i}\left( {\sum\limits_{j = {K_i} + 1}^L {{C_{i,j}}} } \right)} .
\]
The minimization problem is thus equivalent to:
\[
{{\bf{C}}_{opt}}: = \mathop {\arg \min }\limits_{\bf{C}} \sum\limits_{i = 1}^N {\sum\limits_{j = {K_i} + 1}^L {{w_i} C^2_{i,j}} } .
\]
Therefore the mapping required is:
\[
 - \frac{1}{2}{{\bf{y}}^T}{{\bf{W}}_A}{\bf{y}} + {{\bf{b}}_A}^T{\bf{y}} = \sum\limits_{i = 1}^N {\sum\limits_{j = {K_i} + 1}^L {{w_i} C^2_{i,j}} } .
\]
The solution is the following:
\[
{{\bf{b}}_A} = {{\bf{0}}_{NL \times 1}},
\]
\[
{{\bf{W}}_A} =  - 2\left( {\begin{array}{*{20}{c}}
{{{\bf{D}}_1}}&{\bf{0}}& \cdots &{\bf{0}}\\
{\bf{0}}&{{{\bf{D}}_2}}& \cdots &{\bf{0}}\\
{\bf{0}}&{\bf{0}}& \cdots &{{{\bf{D}}_N}}
\end{array}} \right) ,
\]
where
\[
{{\bf{D}}_j} = \left( {\begin{array}{*{20}{c}}
{{{\bf{0}}_{{K_j} \times {K_j}}}}&{{{\bf{0}}_{{K_j} \times \left( L - {K_j} \right) }}}\\
{{{\bf{0}}_{ \left( L - {K_j} \right) \times {K_j}}}}&{{w_j} {{\bf{I}}_{\left( L - {K_j} \right) \times \left( L - {K_j} \right) }}}
\end{array}} \right) \in {\mbox{R}^{L \times L}} .
\]
Having transformed the objective function to a quadratic form, we now turn our attention to doing the same with the two constraints outlined in (\ref{eq:e24}) and (\ref{eq:e25}).

The constraints in (\ref{eq:e24}) can be rewritten as follows:
\[
\forall i:\sum\limits_{j = 1}^L {{C_{i,j}}}  = {x_i} \to \min_\mathbf{C} {\sum\limits_{i = 1}^N {\left( {\left( {\sum\limits_{j = 1}^L {{C_{i,j}}} } \right) - {x_i}} \right)} ^2} .
\]
This will lead to the following equation:
\begin{equation}
 - \frac{1}{2}{{\bf{y}}^T}{{\bf{W}}_B}{\bf{y}} + {{\bf{b}}_B}^T{\bf{y}} = {\sum\limits_{i = 1}^N {\left( {\left( {\sum\limits_{j = 1}^L {{C_{i,j}}} } \right) - {x_i}} \right)} ^2} .
\label{eq:e416}
\end{equation}
The solution of equation (\ref{eq:e416}) is:
\[
{{\bf{b}}_B} = 2\left( {\begin{array}{*{20}{c}}
{{x_{{1_{1 \times L}}}}}&{{x_{{2_{1 \times L}}}}}& \cdots &{{x_{{J_{1 \times L}}}}}
\end{array}} \right) ,
\]
\[
{{\bf{W}}_B} =  - 2\left( {\begin{array}{*{20}{c}}
{{{\bf{1}}_{L \times L}}}&{\bf{0}}& \cdots &{\bf{0}}\\
{\bf{0}}&{{{\bf{1}}_{L \times L}}}& \cdots &{\bf{0}}\\
 \vdots & \vdots & \ddots & \vdots \\
{\bf{0}}&{\bf{0}}& \cdots &{{{\bf{1}}_{L \times L}}}
\end{array}} \right) .
\]
Another constraint ensures that the scheduling does not overload the processing units, described in (\ref{eq:e25}). The required transformation is as follows:
\begin{equation}
\forall j:\sum\limits_{i = 1}^N {{C_{i.j}}}  = V \to \min_\mathbf{C} {\sum\limits_{i = 1}^L {\left( {\left( {\sum\limits_{j = 1}^N {{C_{i,j}}} } \right) - V} \right)} ^2} .
\label{eq:e419}
\end{equation}
Please note the technicality that this constraint may not be relevant for the last few columns where only the remaining jobs have to be scheduled and schedule matrices not exhausting the full capacity should not be penalized. The total length of scheduling can be written as follows:
\[
\tilde L = \left\lceil \dfrac{\sum \limits_{i = 1}^N x_i }{V} \right\rceil ,
\]
\[
L = \max \left( {\tilde L,\mathop {\max }\limits_i {x_i},\mathop {\max }\limits_i {K_i}} \right).
\]
The number of columns where capacity $V$ needs to be fully utilized is the following:
\begin{equation}
M = \max \left( {1,\mathop {\max }\limits_i {x_i} - \hat{L} + 1} \right) .
\label{eq:e421}
\end{equation}
Taking into consideration (\ref{eq:e421}) the mapping of (\ref{eq:e419}) can be described by the following equation:
\[
 - \frac{1}{2}{{\bf{y}}^T}{{\bf{W}}_B}{\bf{y}} + {{\bf{b}}_B}^T{\bf{y}} = {\sum\limits_{i = 1}^M {\left( {\left( {\sum\limits_{j = 1}^N {{C_{i,j}}} } \right) - V} \right)} ^2} .
\]
The solution of this equation is the following:
\[
{{\bf{b}}_C} = \left[ {{{\bf{V}}_{M \times 1}},{{\bf{0}}_{ \left( L - M  \right) \times 1}},{{\bf{V}}_{M \times 1}},{{\bf{0}}_{ \left( L - M \right) \times 1}}, \ldots ,{{\bf{V}}_{M \times 1}},{{\bf{0}}_{ \left( L - M  \right) \times 1}}} \right] ,
\]
\[
{{\bf{W}}_C} =  - 2\left( {\begin{array}{*{20}{c}}
{\bf{D}}&{\bf{D}}& \cdots &{\bf{D}}\\
{\bf{D}}&{\bf{D}}& \cdots &{\bf{D}}\\
 \vdots & \vdots & \ddots & \vdots \\
{\bf{D}}&{\bf{D}}& \cdots &{\bf{D}}
\end{array}} \right) ,
\]
where
\[
{\bf{D}} = \left( {\begin{array}{*{20}{c}}
{{{\bf{I}}_{M \times M}}}&{{{\bf{0}}_{M \times \left( L - M \right) }}}\\
{{{\bf{0}}_{ \left( L - M \right) \times M}}}&{{{\bf{0}}_{ \left( L - M \right) \times \left( L - M \right)}}}
\end{array}} \right) .
\]
We can combine these mappings into the form of equation (\ref{eq:e41}) as follows:
\[
{\bf{W}} = \alpha {{\bf{W}}_A} + \beta {{\bf{W}}_B} + \gamma {{\bf{W}}_C} \in {\mbox{R}^{NL \times NL}},
\]
and
\[
{\bf{b}} = \alpha {{\bf{b}}_A} + \beta {{\bf{b}}_B} + \gamma {{\bf{b}}_C} \in {\mbox{R}^{NL \times 1}}.
\]
Note that the objectives can be controlled with heuristic constants $\alpha ,\beta$ and $\gamma$ in order to strike an  appropriate balance between the weights of different requirements. Having this quadratic form at hand, we can apply the HNN to provide an approximate solution to the optimization problem in polynomial time.

\section{Numerical results}
\label{sec:5NumRes}
In this section, the performance of the HNN approach is investigated and is compared to the performance of other heuristics. 

\subsection{Simulation method}
\label{sec:51Sim}
Each method outlined in Section 3 and the HNN method outlined in Section 4 has been tested by simulation on a large and diverse set of input parameters with the aim to characterize the algorithms empirically on scheduling problems of different size of jobs. The algorithms were implemented in Matlab and tests were run in this simulation environment with randomly generated  parameters such as size of jobs, cutoff times, and weights. 
The size of each job and its cutoff time and corresponding weight are generated as follows:
\begin{equation}
  \mathbf{x}_i = \mbox{random} \left( \left[ 1, c_1 \right] \right),
  \label{eq:e511}
\end{equation}
\begin{equation}
  \mathbf{K}_i = \mathbf{x}_i + \mbox{random} \left( \left[ c_1, 1.5 \cdot c_1 \right] \right),
  \label{eq:e512}  
\end{equation}
\begin{equation}
  \mathbf{w}_i = \mbox{random} \left( \left[ 1, c_2 \right] \right),
  \label{eq:e513}
\end{equation}
where $\mbox{random} \left( \Theta \right)$ produces a uniformly distributed random integer value in range $\Theta$.
In our simulations the constant $c_1$ equals $10$ and $c_2$ equals $5$. The number of processors ($V$) is determined as follows:
\begin{equation}
  V = 0.25 \cdot J.
    \label{eq:e514}
\end{equation}
The problem is expected to be solved without tardiness when the capacity is $V = J$. Therefore (\ref{eq:e514}) ensures that there is a high likelihood of tardiness associated with the generated problem.

For each problem size, $100$ different problems were generated randomly, using (\ref{eq:e511})-(\ref{eq:e514}) and the results the methods were compared in each case.

The random method was repeated $1000$ times for each problem and the best solution was used. Furthermore, the HNN method was repeated $1000$ times for each problem with different random starting points and the best solution was used. In addition, the heuristical parameters $\alpha ,\beta$ and $\gamma$ were adjusted between the simulations in order to provide a good balance between optimizing the objective function, but also meeting the required constraints to produce a valid scheduling matrix.
To adjust the heuristical parameters we used Algorithm \ref{alg:a1}.

\begin{algorithm}
\begin{algorithmic}
\REQUIRE $\mathbf{x}, \mathbf{K}, V, e$
\STATE $\alpha \gets 0.1, \beta \gets 5, \gamma \gets 5$
\STATE $i \gets 0$
\REPEAT
	\STATE $i \gets i + 1$
	\STATE $\mathbf{C}_i \gets \mbox{HNN} \left( \mathbf{x}, \mathbf{K}, V, \alpha, \beta, \gamma \right)$
	\STATE $\alpha \gets \alpha + 0.01$
\UNTIL{$\mbox{errors} \left( \mathbf{C}_i \right) \leq e$}
\FOR{$k = 1 \to i$}
	\STATE $\mathbf{C}_k \gets \mbox{correct} \left(\mathbf{C}_k\right)$
	\STATE $\mathbf{T}_k \gets \mbox{calculateTWT} \left( \mathbf{C}_k \right)$
\ENDFOR
\RETURN $\min \left( \mathbf{T} \right)$
\caption{Algorithm for adjusting the heuristical parameters \label{alg:a1}}
\end{algorithmic}
\end{algorithm}

\begin{algorithm}
\begin{algorithmic}
\REQUIRE $\mathbf{x}, \mathbf{w}, \mathbf{K}, V, \mathbf{C}$
\FOR{$k = 1 \to L$}
	\WHILE {$\sum_{i=1}^N \mathbf{C}_{i,k} > V$}
		\STATE Remove $1$ from row $j$, where $j$  is the row in column $k$ with minimal weight that has $C_{j,k} = 1$
	\ENDWHILE
\ENDFOR
\FOR{$k = 1 \to N$}
	\WHILE {$\sum_{i=1}^L \mathbf{C}_{k,i} > \mathbf x_k$}
		\STATE Remove $1$ from row $k$ from the column $l$, where $l$ is the righternmost column in row $k$ such that $C_{k,l}=1$
	\ENDWHILE
	\WHILE {$\sum_{i=1}^L \mathbf{C}_{k,i} < \mathbf x_k$}
		\STATE Add $1$ to row $k$ in column $l$ where $l$ is the lefternmost column where $1$ can be added without violating the capacity constraint $V$.
	\ENDWHILE	
\ENDFOR
\RETURN $\mathbf{C}$
\caption{\label{alg:a2}Correction algorithm for the scheduling matrix produced by the HNN}
\end{algorithmic}
\end{algorithm}

Even so, in some cases the HNN is unable to provide a valid solution, because the prescribed constraints on job sizes and capacity are violated. Therefore, we introduced an error correction function (see Algorithm \ref{alg:a2}) in order to cut and replace the unnecessary 1-s in the scheduling matrix. In our simulations parameter $e$ equals $5$. All of the results presented in the following sections concern valid schedules meeting the required constraints.

\subsection{Average total weighted tardiness of the different methods}
\label{sec:52TWT}

The first simulation compares the average total weighted tardiness provided by the algorithms for different problem sizes.

\begin{figure*}[!htb]
\includegraphics[width=130mm]{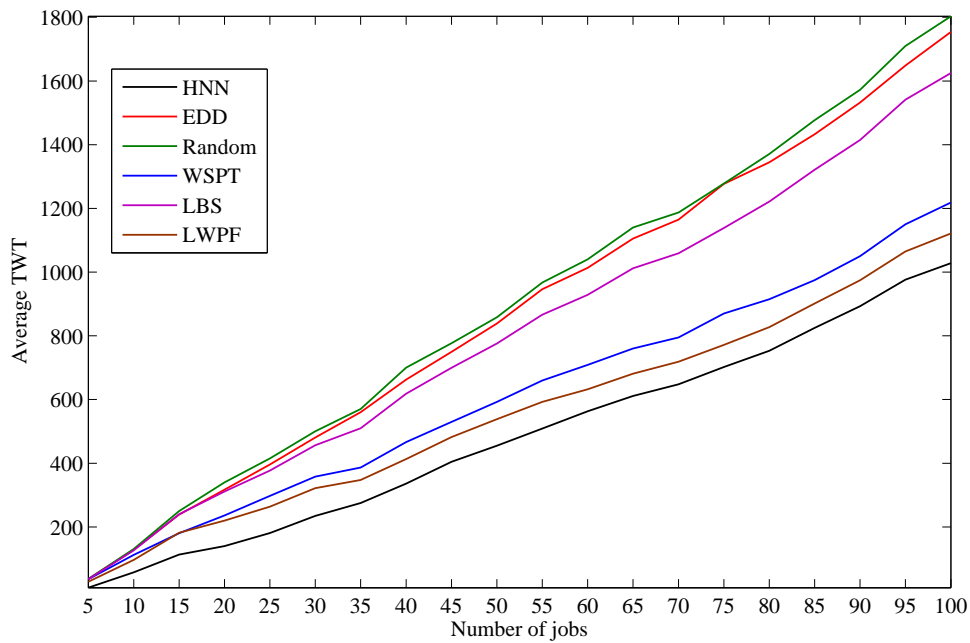}
\caption{Average TWT produced by each heuristic over randomly generated problems depicted as a function of the number of jobs in the problem.}
\label{fig:1}      
\end{figure*}
Figure \ref{fig:1} demonstrates that the best solution achieved by the HNN produces better average TWT for all problem sizes than any of the other heuristics. The performances of the EDD, LBS and random methods are, on average, worse than the WSPT and LWPF heuristics and the HNN for all problem sizes. This shows that consideration of the weights in the optimization problem is critical to reach near-optimal TWT schedules. It is also clearly visible that the HNN consistently outperforms all other methods for all problem sizes. Note also that the simple LWPF heuristic, proposed by us, outperforms the WSPT heuristic widely used as a benchmark in the literature.

\subsection{Detailed comparison of the HNN performance versus other heuristics}
\label{sec:54DHNN}

\begin{figure*}[!htb]
\includegraphics[width=130mm]{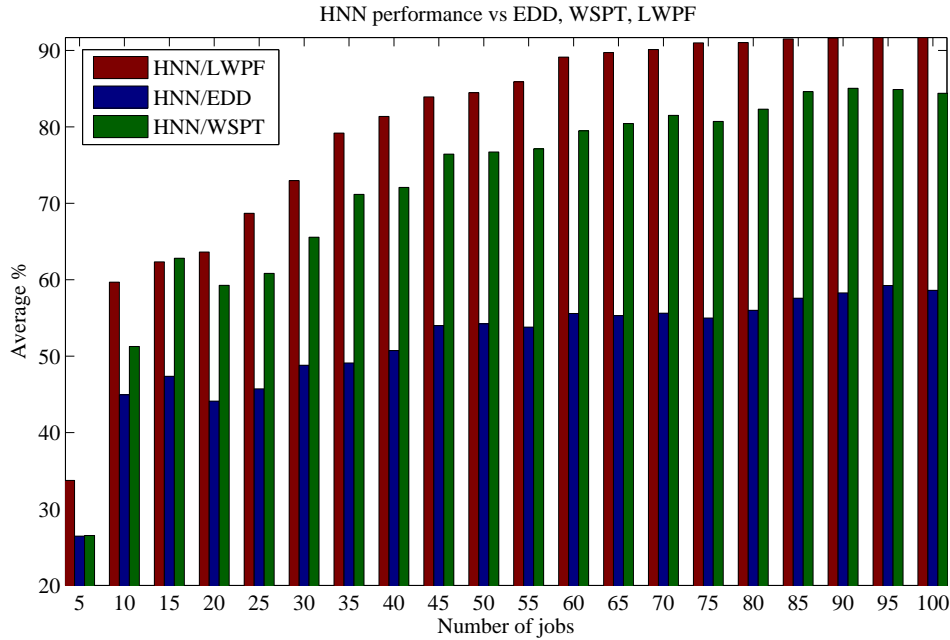}
\caption{Ratio of average TWT produced by the HNN versus EDD, WSPT and LWPF as a function of the problem size}
\label{fig:2}      
\end{figure*}

In this section, we quantify how much better the solution provided by the HNN is, in comparison to the EDD, WSPT and LWPF algorithms. Figure \ref{fig:2} depicts the ratio of average TWT produced by the HNN method versus the other heuristics. We conclude that, on average, the best solution of the HNN heavily outperforms the traditional solutions: the total weighted tardiness of the best HNN solution is $25\%-56\%$ of the EDD, $25\%-84\%$ of the WSPT, and $34\%-91\%$ of the LWPF. As the problem size increases, the WSPT and LWPF heuristics produce solutions which are closer to the HNN.

In order to verify that the HNN consistently outperforms the other heuristics on a wide spectrum of problems, not just on a few selected problems for each given problem size, we ran a more detailed experiment. In this case, we investigated $500$ randomly generated problems for each problem size and computed the percentage of times the HNN produced a better solution than LWPF, the next best method. Table \ref{tab:1} shows the result which quite convincingly proves the general applicability of the HNN to different problem types. Across the spectrum of problem domain, the ratio is higher than $98.5\%$ for all investigated job sizes. 
\begin{table}[!h]
\center
\caption{Percentage of problems in which the HNN provides an improved solution over the next best heuristic, LWPF}
\label{tab:1}   
\begin{tabular}{l|ccccccccc}
\bf{Job size} & \bf 5 & \bf 10 & \bf 20 & \bf 25 &  \bf 50 & \bf 75 & \bf 100  \\ \hline
 & 99.9\% & 100\% & 99.5\% & 99.2\% & 99.3\% & 98.6\% & 98.8\% \\ \end{tabular}
\end{table}

\subsection{Runtime characteristics of the investigated algorithms}
\label{sec:55Runtime}
In this section, we summarize the runtime characteristics of the introduced algorithms. Table \ref{tab:2} contains the theoretical order of convergence of the algorithms as a function of the length of the input parameters.

\begin{table}[!h]
\center
\caption{Theoretical order of convergence of the investigated algorithms}
\label{tab:2}
\begin{tabular}{l|l|l}
\textbf{Algorithm} & \textbf{Order of convergence}            & \textbf{Reference} \\ \hline
Random strategy   & $\textit{O} \left(L \cdot N \right)$     & \cite{Laszlo2011} \\
EDD, WSPT, LWPF   & $\textit{O} \left(L \cdot N \right)$     & \cite{Maheswaran2004} \\
LBS               & $\textit{O} \left(L \cdot N^2 \right)$   & \cite{Laszlo2011} \\ 
HNN               & $\textit{O} \left(L^2 \cdot N^2 \right)$ & \cite{Hopfield1982}, \cite{Haykin2008} \\
Exhaustive search & $\textit{O} \left(2^{L \cdot N} \right)$ &
\end{tabular}
\end{table}

Note that this table shows the theoretical runtime for a single run of the specified algorithm, whilst we need to run a constant number of the HNN and random iterations. On the other hand, these independent runs can be executed in parallel on several computers, multicore machines or even GPU-based architectures, so the overall runtime should not be significantly different due to the need to run multiple iterations.

\begin{figure*}[!h]
\includegraphics[width=130mm]{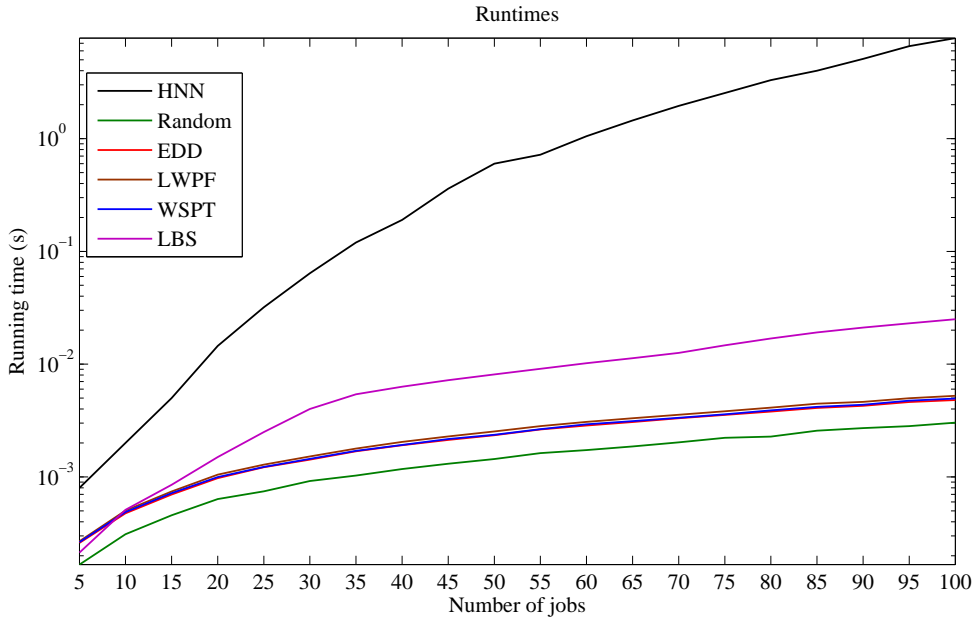}
\caption{The average runtime of each algorithm per iteration (including the iterative adjustment of the heuristic parameters as per Algorithm 1 for HNN), depicted as a function of the number of jobs. Note that the vertical axis is using a logarithmic scale}
\label{fig:3}      
\end{figure*}

Figure \ref{fig:3} depicts the empirical runtime comparison of the different heuristics which confirm the theoretical limits. Although the HNN is the slowest method of the investigated heuristics, in practice we see that near optimal solution of a $100$ job problem takes only around 6 seconds / iteration (including the iterative adjustment of the heuristic parameters as per Algorithm \ref{alg:a1}) on Core i7@3Ghz processor, running non-optimized Matlab code, which is very promising as far as its practical applicability in real life applications is concerned. With further optimization and parallelization on multicore machines or GPGPU technology, the HNN method provides better quality solutions than the other heuristics within acceptable runtimes.

\section{Conclusions and directions of future research}
\label{sec:6Conlusion}

In this paper, we studied the NP-hard problem of scheduling jobs with given relative priorities (weights) and deadlines on identical machines, minimizing the TWT measure. We developed a novel heuristic approach, utilizing quadratic programming and the Hopfield neural network and we showed that, in general, it outperforms the EDD and WSPT methods. Furthermore, we have shown that our approach can be applied in real-time settings for small problem sizes and possesses scalability features which are acceptable for real world applications.

More formal methods for the selection of alpha, beta and gamma parameters could be investigated in the future to further improve performance. Another idea to explore is whether a more intelligent selection of the initial point for the HNN algorithm (eg. the result of the LWPF or WSPT heuristics) would improve the algorithms performance over the random starting point which was used in our tests. Finally, the performance of the HNN algorithm needs to be compared to more complex heuristic methods.

\bigskip
\rightline{\emph{Received: March 5, 2012 {\tiny \raisebox{2pt}{$\bullet$\!}} Revised: April 16, 2012}} 
 
\end{document}